\documentclass{article}

\usepackage{microtype}
\usepackage{graphicx}
\usepackage{subfigure}
\usepackage{booktabs} % for professional tables

\usepackage{times}
\usepackage{epsfig}
\usepackage{amsmath}
\usepackage{amssymb}
\usepackage{multirow}
\usepackage{array}
\usepackage{makecell}
\usepackage{bm}
\usepackage{enumerate}
\usepackage[T1]{fontenc}
\usepackage{algorithm}
\usepackage{algorithmic}
\newtheorem{theorem}{Theorem}
\usepackage[breaklinks=true,bookmarks=false]{hyperref}

\usepackage[preprint]{neurips_2020}

\title{Reducing Conservativeness Oriented Offline Reinforcement Learning}

\author{
Hongchang Zhang, Jianzhun Shao, Yuhang Jiang, Shuncheng He, Xiangyang Ji\\
Department of Automation\\
Tsinghua University\\
\{hc-zhang19,sjz18, jiangyh19,hesc16\}@mails.tsinghua.edu.cn, xyji@tsinghua.edu.cn
}

\begin{document}

\maketitle

\begin{abstract}
In offline reinforcement learning, a policy learns to maximize cumulative rewards with a fixed collection of data. Towards conservative strategy, current methods choose to regularize the behavior policy or learn a lower bound of the value function. However, exorbitant conservation tends to impair the policy's generalization ability and degrade its performance, especially for the mixed datasets. In this paper, we propose the method of reducing conservativeness oriented reinforcement learning. On the one hand, the policy is trained to pay more attention to the minority samples in the static dataset to address the data imbalance problem. On the other hand, we give a tighter lower bound of value function than previous methods to discover potential optimal actions. Consequently, our proposed method is able to tackle the skewed distribution of the provided dataset and derive a value function closer to the expected value function. Experimental results demonstrate that our proposed method outperforms the state-of-the-art methods in D4RL offline reinforcement learning evaluation tasks and our own designed mixed datasets.
\end{abstract}

\section{Introduction}
Reinforcement learning has gained unprecedented attention in recent years, benefiting from the powerful representation and expressive capability of deep neural networks \cite{mnih2015human, schulman2017proximal}. 
However, contemporary deep reinforcement learning algorithms generally require millions of samples to achieve expert-level performance. 
In many real-world applications, such as healthcare and self-driving, trial-and-error collecting samples is costly, time-consuming, and even unsafe. 
It is becoming a major bottleneck in deep reinforcement learning for the employment in these scenarios \cite{dulac2019challenges}.

By utilizing the logged dataset, offline reinforcement learning is one promising solution for the above issues.
The policy can be learned without further interaction with the environment. 
The dataset could be generated by other agents, other algorithms, and even other tasks. 
For instance, the Atari videos on Youtube can be employed to train the policy to imitate human gameplay\cite{aytar2018playing}. 
Besides, without the chance to interact with the environment, the algorithm cannot perform hazardous options throughout the training period.
However, current reinforcement learning methods easily fail to find the optimal policy due to limited and narrow datasets.
\cite{fujimoto2019off} showed offline reinforcement learning suffers from the extrapolation error, which is caused by the optimistic estimation of value target of unseen state-action pairs. 
Numerous researchers attempt to reduce the extrapolation error by learning the policy in the constrained offline action space. \cite{kumar2019stabilizing, wu2019behavior} minimized the divergence between the trained policy and the offline policy. \cite{kumar2020conservative} proposed Conservative Q-Learning (CQL) to learn a lower bound of the expected value function and update the policy according to the conservative value function.

Although these methods use conservative strategies to behave close to the offline distribution, exorbitant conservativeness might impair the policy's generalization ability, especially in high-dimensional settings. 
For CQL, learning a lower bound of the value function can suppress the effect of out-of-distribution samples.
However, the discrepancy between the lower bound and the real value function might pose challenges to discover optimal actions. 
Especially for non-expert datasets, optimal policy might deviate far from the offline policy. The conservative lower bound concerns staying in the offline distribution and easily fails to derive the optimal policy.

Prior conservative methods always penalize samples with high uncertainty. Note that the out-of-distribution samples and those samples with small occurrences in the dataset both generally have high uncertainty. However, they are not supposed to be treated equally, especially for the skewed dataset where valuable samples are relatively sparse. 
Training the policy conservatively in terms of a imbalanced distribution will undermine the agent's performance. 
The data imbalance always appears in supervised learning, where one or more categories occupy relatively small proportions of the training data against that of the others. 
In offline reinforcement learning, there is always a disproportionate percentage of state-action pairs for each component in policy mixtures. 
The previous methods generally neglect data imbalance problem in offline datasets. 
Although these methods achieve remarkable results in most D4RL evaluation benchmarks \cite{fu2020d4rl}, they fail to perform well as expected in the mixed datasets. Mixed datasets are introduced by various policies, which are prevalent in many applications. 
For these tasks, it is crucial to reduce the conservativeness by paying more attention to the sparse samples.

In this paper, we propose reducing conservativeness oriented offline reinforcement learning to tackle the issues.
On the one hand, we expect to treat the majority and minority samples equally to handle the data imbalance problem.
We first provide the samples in the dataset with pseudo class labels by clustering, and then employ resampling and reweighting to train the samples according to each transition's class frequency.
In general, the samples with rare occurrences are assumed to have high uncertainty. Enhancing the weight of these samples in training might introduce more uncertainty and enlarge the extrapolation error. 
Third, to reduce the extrapolation error resulting from uncertainty preference, we choose to penalize the out-of-distribution samples based on their uncertainty. 
In this manner, the minority samples are supposed to be attached with more importance while the trust-region of these samples will be relatively cramped. 
On the other hand, we introduce a tighter lower bound into the value function optimization process, which is able to reduce the conservativeness further. 
The tighter lower bound is expected to explore the potential optimal actions for the agents, and is crucial when presented with non-expert datasets.
We evaluate our proposed method on D4RL Mujoco tasks and our own designed mixed dataset.
The experimental results demonstrate that our proposed method outperforms and achieves the competitive performance compared with the state-of-the-art methods on the most tasks.

\section{Related Work}
The offline reinforcement learning methods \cite{lange2012batch} focus on training a policy given a fixed dataset. 
They have been applied in various applications, such as healthcare \cite{gottesman2018evaluating,wang2018supervised}, recommendation systems \cite{strehl2010learning,swaminathan2015batch,covington2016deep,chen2019top}, dialogue systems\cite{zhou2017end}, and autonomous driving \cite{sallab2017deep}.
In these settings, standard off-policy methods \cite{mnih2015human,lillicrap2015continuous} 
generally fail and exhibit undesirable performance due to their sensitiveness to the dataset distribution \cite{fujimoto2019off}. 
The problem has been studied in approximate dynamic programming \cite{bertsekas1995neuro, farahmand2010error, munos2003error, scherrer2015approximate}. 
They ascribed it to the errors arising from distribution shift and function approximation.
Recently, \cite{van2018deep} stated that the temporal difference algorithms would diverge when represented with function approximators and trained under off-policy data.

Some methods have been proposed to deal with the issue by restraining the policy distribution conservatively. 
\cite{fujimoto2019off} proposed Batch-Constrained Reinforcement Learning (BCQ) to select actions close to the offline actions when formulating Bellman target.
\cite{kumar2019stabilizing} presented a bootstrapping error accumulation reduction (BEAR). They minimized Maximum Mean Discrepancy (MMD) between the trained and offline policies to alleviate the Bellman update's error propagation.
\cite{jaques2019way} utilized KL-divergence with a regularization weight to penalize the policy.
\cite{wu2019behavior} introduced value penalty and policy penalty to regularize the learned policy towards the behavior policy. 
These methods tried to penalize the out-of-distribution actions and force the policy to behave in the dataset's action space. However, these methods ignore the data imbalance problem in the logged dataset. In an optimistic view, we focus on treating the minority classes and the majority classes equally. 
Our propsoed method is closely related to CQL\cite{kumar2020conservative}, which learned a lower bound for the policy's value function by penalizing the out-of-distribution state-action pairs and maximizing the in-distribution samples' Q-value. 
Differently, we encourage the violation from the offline policy by estimating a tighter lower bound for the value function, which is helpful to non-expert tasks. 
Similarly, \cite{agarwal2019striving} used an optimistic strategy and introduced a value function ensemble to stabilize the Q-function updating. Although our method is proposed in terms of reducing the conservativeness, it is essentially a conservative method.

Previous methods tend to penalize samples with high uncertainty. 
\cite{yu2020mopo} trained an auxiliary dynamics model to produce imaginary samples to augment the dataset. 
They reduced a reward for the generated samples according to their uncertainty. 
\cite{kidambi2020morel} learned a pessimistic Markov Decision Process (MDP) using the offline dataset and as well as a near-optimal policy based on this pessimistic MDP. 
By employing an unknown state-action detector, they used a model-based method to train the policy.
However, our proposed method encourages the policy to utilize samples with high uncertainty instead. 
Furthermore, we impose strict constraints on these sparse samples to reduce extrapolation error.

Current methods of tackling data imbalance can be divided into two categories, i.e., resampling and cost-sensitive learning. 
Resampling means over-sampling \cite{shen2016relay} the minority classes and under-sampling\cite{he2009learning} the frequent classes. 
Over-sampling could result in the over-fitting of minority classes, while under-sampling might ignore valuable information in the majority classes. 
Cost-sensitive learning \cite{wang2017learning} assigns different weights for different samples. 
Most methods reweight samples according to the inverse of the class frequency. 
\cite{cui2019class} proposed to reweight using the effective number of the classes. 
In reinforcement learning, Monte Carlo dropout \cite{srivastava2014dropout} and random network distillation \cite{burda2018exploration} are common methods to estimate the frequency of each sample. 
Since offline datasets do not offer class labels for samples, we introduce pseudo labels to the samples using unsupervised learning methods. 
In this manner, our proposed method is able to perform resampling and reweighting efficiently for different clusters.

\section{Background}
In general, a reinforcement learning problem could be formulated as a Markov decision process $(S, A, P, R,\gamma)$, with state space $S$, action space $A$, transition probability matrix $P$, reward function $R$, and discount factor $\gamma$ \cite{sutton2018reinforcement}. 
Each term of $P$ denotes the probability of arriving at the next state $s'$ when selecting $a$ at state $s$. 
At each time step, an agent is located at $s$ and chooses action $a$. 
The agent would arrive at a next state $s'$ and receive a reward $r$ from the environment. 
The objective of reinforcement learning is to train a policy $\pi(a|s)$ to maximize accumulative rewards in one episode as:
\begin{equation}
\label{eq1:rl}
\max_{\pi} \mathbb{E}[\sum_{t=0}  \gamma^t r_t].
\end{equation}
When the agent performs an action $a$ at state $s$ according to a policy $\pi$, the expectation of the cumulative rewards is defined as Q-value function: $Q(s,a) = \mathbb{E}_{\pi}[\sum_{t=0} \gamma^t r_t | s,a]$. 
And the value function is defined as $V(s)=\mathbb{E}_{\pi}[Q(s,a)]$. 
The Q-learning algorithms update the Q-value function according to Bellman operator $\mathcal{T}$ as:
\begin{equation}
\label{eq2:bellman}
\mathcal{T}Q(s,a) := \mathbb{E}[r +\gamma \max_{a'} Q(s',a')]
\end{equation}

Compared with Q-learning methods, actor-critic methods employ a policy $\pi(a|s)$ to sample actions. The update rule of Q-value function evaluation is $\mathcal{T}^\pi Q = r + \gamma P^{\pi}Q$, where $P^{\pi}Q$ is the expectation of Q-value with respective to transition matrix $P$ and policy $\pi$.
Since it is impossible to enumerate all states and actions, the Q-value function is generally updated using an empirical Bellman operator $\hat{T}$. 
Recently, \cite{haarnoja2018soft} employed deep neural networks to approximate the actual Q-value function and the policy. 
On the basis of the neural network, the Q-value loss function is then defined as: 
\begin{equation}
\label{eq3:sac}
E_{s,a,r,s'\sim \mathcal{D}}(Q(s,a;\theta) - (r + \gamma(Q(s',\pi(a'|s';\omega);\theta))) ^ 2,
\end{equation}
where $\mathcal{D}$ is a replay buffer containing the collected samples.
In Eq. \ref{eq3:sac}, $\theta$ is the parameter of the Q-value network and $\omega$ is the parameter of the policy. 
In this manner, the policy can be updated as:
\begin{equation}
\label{eq4:policy}
\max_{\omega} Q(s,\pi(a|s;\omega);\theta).
\end{equation}

Offline reinforcement learning suffers from the out-of-distribution challenge.
The standard reinforcement learning algorithms might derive a policy that generates out-of-distribution actions when applied in this setting. 
They have no chance to correct the real Q-value of these actions. 
Recent works focus on punishing the out-of-distribution actions. 
\cite{kumar2020conservative} proposed the CQL method by maximizing the value of in-distribution state-action pairs and minimizing the value of out-of-distribution state-action pairs. 
The Q-network is updated as:
\begin{equation}
\begin{aligned} & \min_Q  \alpha(\mathbb{E}_{s \sim \mathcal{D}, a \sim \pi(\cdot \mid s)}[Q(s, a)] -\mathbb{E}_{s \sim \mathcal{D}, a \sim \pi_\beta(a|s)}[Q(s, a)]) \\
& + \mathbb{E}_{s, a, s^{\prime} \sim \mathcal{D}}\left[\left(Q(s, a)-\hat{\mathcal{T}}^\pi \hat{Q}(s, a)\right)^{2}\right].
\end{aligned}
\label{eq5:cql}
\end{equation}
In Eq. \ref{eq5:cql}, $\alpha$ is a hyperparameter for controlling the degree of conservativeness with respect to the Bellman update, $\hat{Q}$ denotes the empirical Q-value function, and $\pi_\beta$ is used to denote the offline policy distribution.

\section{Data Imbalance}
In real-world applications, a large amount of data is generally produced by the skewed distributions. 
The samples of some classes outnumber those of other classes.
In healthcare applications, severe events like hematorrhea are rare but important. 
In advertisement systems, the number of valuable purchases is much smaller than that of clicks. 
In this section, we present the discussion of the data imbalance in offline reinforcement learning.

The data imbalance phenomenon exists in the dataset generated by a single policy.
Without loss of generality, we assume the transition model of the environment is stochastic in the single policy scenario. 
A suboptimal stochastic policy might produce a small portion of expert-level trajectories and a large portion of trajectories with poor performances. 
During the training process, those suboptimal samples dominate over the sparse samples of high quality. 
In standard reinforcement learning, the temporal difference method samples the experiences uniformly. 
The expert-level trajectories would be sampled inefficiently, which will impair the reward propagation and undermine the training process. 
Moreover, the weights of sparse samples might lead to large variances for off-policy evaluation.

Mixed policies are characterized by data imbalance and are ubiquitous in real-world applications.
In autonomous driving, samples are produced by different drivers with different driving inclinations. 
Similarly, in reinforcement learning settings, the agents interact with the environment using the new updated policy at each time step. The collected samples produced by different policies lead to a mixed dataset.
The induced skewed dataset consists of samples produced by different policies with different proportions. 
The large proportion of the samples dominates the training process of the policy compared with that of the small proportion of the samples.
In the worst cases, the small proportion of samples from high-quality policies is hard to be utilized.

\begin{figure}[th] 
\includegraphics[width=1.0\linewidth]{ 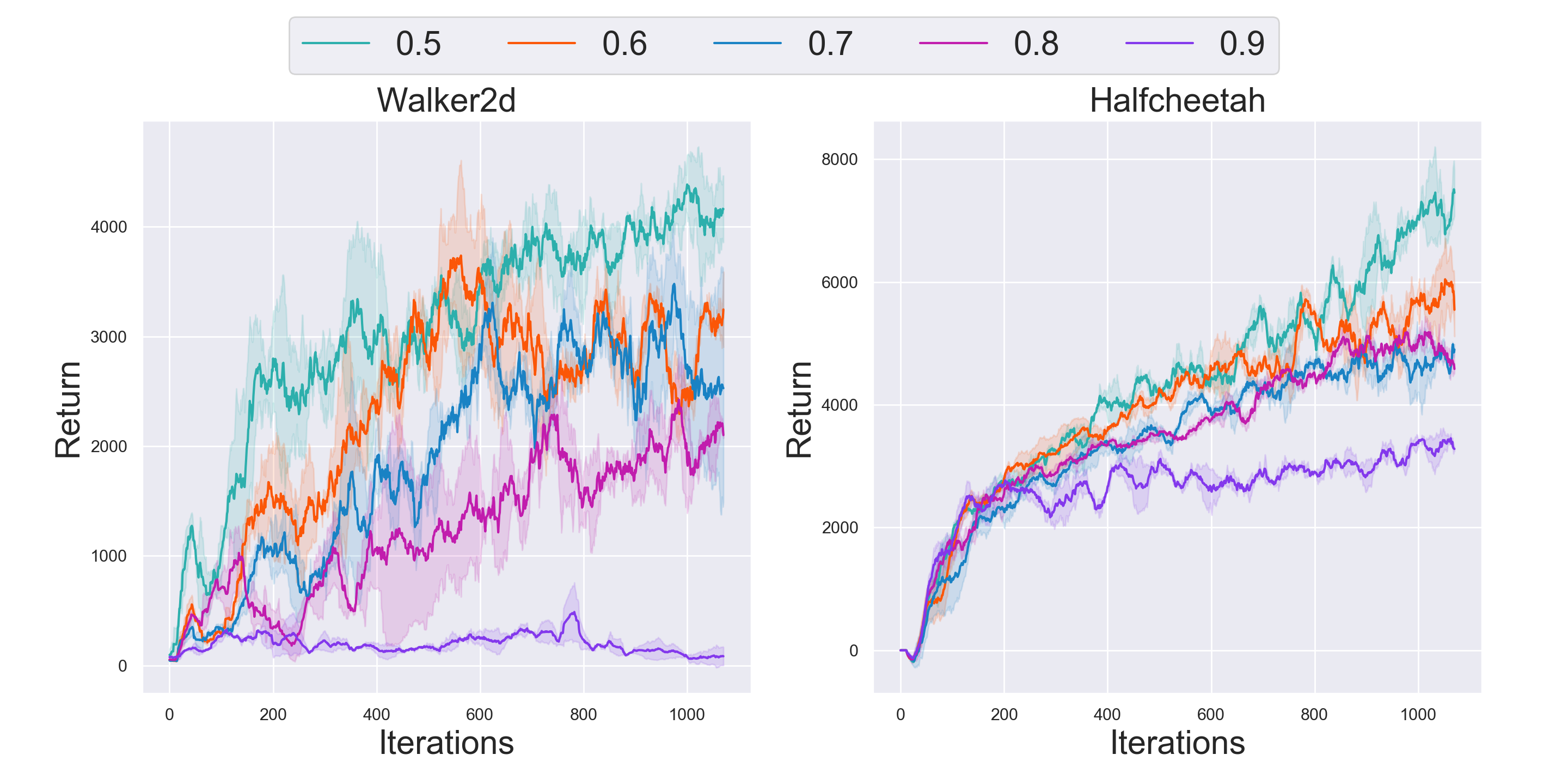}
\caption{Results of CQL on ``Walker2d-random-expert'' and ``Halfcheetah-random-expert'' tasks with different proportions of random samples. Note that as the random proportion increases, the performance of CQL decreases.}
\label{fig2:imbalance}
\end{figure}

We examine the performance of the offline algorithm using D4RL benchmarks on the skewed datasets.
We study the Mujoco tasks Walker2d and Halfcheetah by mixing the ``random'' and ``expert'' dataset with different ratios for evaluation. 
We create five datasets and all the datasets consist of $10^6$ samples for each task.
And the proportions of random samples are set as $0.5$, $0.6$, $0.7$, $0.8$, and $0.9$ respectively.
Accordingly, the proportions of expert-level samples decline. 
We train a conservative Q-learning (CQL) agent \cite{kumar2020conservative} for $10^6$ updates and run four seeds for each experiment.
The results for these tasks are shown in Fig. \ref{fig2:imbalance}. 
As demonstrated in Fig. \ref{fig2:imbalance}, the performance of the trained policy decreases as the random sample proportion increases. 
As the above discussion, the current offline reinforcement learning methods struggle to handle the data imbalance problem.  

\section{Lower Bound of the Value Function}
The CQL proposed to learn a lower bound of the expected value function $V(s)$ by minimizing the Q-value of out-of-distribution state-action pairs and maximizing the Q-value of in-distribution state-action pairs. 
In the setting without approximation error, the gap between the value function derived by CQL and expected value function is the Pearson $\chi^{2}$ divergence between $\pi_{\beta}(a|s)$ and $\pi(a|s)$. 
The gap decreases to $0$ when $\pi(a|s) = \pi_{\beta}(a|s), \forall a$. 
We could obtain a baseline $\delta(s)$ which is the minimum of Pearson $\chi^{2}$ divergence for $s$ and a threshold as:
\begin{equation}
\delta(s)= \min\{\sum_a  \frac{(\pi(a|s) - \pi_\beta(a|s))^2}{\pi_\beta(a|s)}, \Delta \},
\label{eq5.1}
\end{equation}
where $\Delta$ is the tolerance threshold of the violation of the trained policy against the offline policy $\pi_{\beta}(a|s)$.
When $\Delta$ decreases, the algorithm tends to behave more conservatively. 
When $\Delta$ increases, the policy is allowed to deviate further from the dataset. 
In this manner, the policy is able to preserve the opportunity to balance stability and generalization ability. 
The objective of our proposed method is then defined as:
\begin{equation}
\begin{aligned}   \min_{Q} \quad& \alpha \cdot(\mathbb{E}_{s \sim \mathcal{D}, a \sim \pi(a \mid s)}[Q(s, a)] \\ & - \mathbb{E}_{s \sim \mathcal{D}, a \sim \hat{\pi}_{\beta}(a \mid s)}[(1+\delta(s))Q(s, a)]) \\ & +\frac{1}{2} \mathbb{E}_{s,a, s^{\prime} \sim \mathcal{D}}\left[\left(Q(s, a)-\hat{\mathcal{T}}^\pi \hat{Q}(s, a)\right)^{2}\right]. \end{aligned}
\label{eq5.2}
\end{equation}

We assume that $C$ is a constant dependent on the reward function and transition matrix, $R_{\max }=\max|r(s,a)|$ and $|\mathcal{D}|$ denotes a vector of counts for all states. 
We show that the empirical value function $\hat{V}^{\pi}(s)$ is the lower-bound of the actual value function.

\begin{theorem}
For any policy $\pi(a|s)$ whose Q-value function is updated as Eq. \ref{eq5.2}, the empirical value function $\hat{V}^\pi (s)$ is a lower bound of the expected value function $V^\pi (s)$ as:
$
\begin{aligned}
    \hat{V}^{\pi}(s) \leq  & V^{\pi}(s)-\alpha\left[\left(I-\gamma P^{\pi}\right)^{-1} \mathbb{E}_{\pi}\left[\frac{\pi}{\hat{\pi}_{\beta}}-(1+\delta)\right]\right](s) \\ & +\left[\left(I-\gamma P^{\pi}\right)^{-1} \frac{C R_{\max }}{(1-\gamma) \sqrt{|\mathcal{D}|}}\right] , \forall s \in \mathcal{D}
\end{aligned}
$
\end{theorem} 
Thus, if $\alpha>\frac{C R_{\max }}{1-\gamma} \cdot \max_{s \in \mathcal{D}} \frac{1}{\mid \sqrt{|\mathcal{D}(s)|}} \cdot \left[\sum_{a} \pi(a \mid s) (\frac{\pi(a \mid s)}{\hat{\pi}_{\beta}(a \mid s)}- (1+\delta(s))) \right] ^{-1}, 
\forall s \in \mathcal{D}$, we have $\hat{V}^{\pi}(s) \leq V^{\pi}(s)$ with high probability. 
When $\hat{\mathcal{T}}^{\pi}=\mathcal{T}^{\pi}$, then any $\alpha>0$  guarantees $\hat{V}^{\pi}(s) \leq V^{\pi}(s), \forall s \in \mathcal{D}$.

\section{Method}
In this section, we introduce our proposed reducing conservativeness oriented offline reinforcement learning. 
We first perform clustering on the dataset and providing the samples with pseudo labels. 
We then employ resampling and reweighting to associate the minor classes with greater importance.
In order to reduce extrapolation error resulting from high uncertainty, we impose regularization constraints on the minority classes. 
Finally, we introduce our proposed tighter lower bound to the value function update.

\subsection{Clustering}
In supervised learning, resampling and reweighting are two typical data preprocessing methods for imbalanced data. 
However, these two methods can not be directly applied to offline reinforcement learning. 
For the data imbalance of a single policy, the majority and minority of classes are latent and should be justified explicitly. 
For the data imbalance of mixed policies, the number of policies and their corresponding samples are generally inaccessible. 
Even if the proportions of the samples are provided, some policies are more similar in state-action space while others are relatively distant but might contain more information.  
As a result, treating these classes equally tends to bring additional data imbalance issues.  
In our proposed method, we create pseudo labels by classifying these samples with an unsupervised learning method.

Given a dataset with $n$ samples, we perform clustering to partition $n$ samples into $k$ clusters. 
For the Mujoco control suite, the dimension of the feature is no more than $30$. 
K-means clustering shows satisfactory performance in this setting.
The advanced clustering methods such as \cite{jiang2016variational} can also be employed.
However, during the experiment, these methods do not gain obvious improvement compared with that of the k-means clustering in this scenario.
We concatenate $(s,a,r)$ to form a new vector and normalize the dataset to a Gaussian distribution with zero mean and unit variance. 
Then the rescaled dataset is clustered using Euclidean distance metric.
K-means is sensitive to initialization and often converges at a local optimum.
We choose the first cluster center uniformly at random from the dataset, and choose the subsequent cluster centers from remaining data samples with probability proportional to their distances from previous closest existing cluster centers.

\subsection{Resampling}
After clustering, we could associate each sample with a pseudo label $l_j$. 
We use $n_{l_j}$ to denote the number of samples of cluster $l_j$.
And during the training process, we classify each transition according to its cluster, and resample samples as:
\begin{equation}
p(s_j,a_j,r_j,s'_j) = \frac{\beta} {n_{l_j}^\eta}.
\label{eq6.1}
\end{equation}
In Eq. \ref{eq6.1}, $\beta$ is a parameter to normalize the resampling distribution, and $\eta$ is a hyperparameter to control the degree of resampling. 
When $\eta=0$, the resampling is reduced to the conventional sampling. 
As $\eta$ increases, the resampling could select sparse samples more frequently.
It should be noted that resampling might result in the overfitting of minority classes. 
It also might discard valuable examples that are important for reward propagation in reinforcement learning. 
We therefore seek a complementary method to handle its shortcomings.

\subsection{Reweighting}
Compared with resampling, reweighting is an alternative method that samples uniformly from the static dataset against data imbalance issue. 
Given transition $(s_j,a_j,r_j,s_j')$, we reweight the loss function in accordance with the inverse of its class frequency. 
With the slight abuse of notation, the reweighted loss function of Q-value is defined as:
\begin{equation}
E_{(s_j,a_j,r_j,s_j')\sim \mathcal{D}} \frac{\beta} {n_{l_j}^\eta} (Q(s_j,a_j) - (r_j + \gamma Q(s_j',\pi(a_j'|s_j'))) ^ 2.
\end{equation}
Assigning sample weights inversely proportionally to the class frequency leads to poor performance in large-scale datasets. 
Since samples share the high-level mutual information, the information gain would decrease as more samples are collected. 
Similar to \cite{cui2019class}, we take advantage of the effective number of each class.  
The weight for transition $\{s_j,a_j,r_j,s_j\}$ is defined as:
\begin{equation}
w(s_j,a_j) = \frac{\beta * (1-\lambda)} {1 - \lambda^{n_{l_j}}},
\label{eq6.3}
\end{equation}
where $\lambda$ is a hyperparameter controlling the degree of reweighting. 
When $\eta=0$, each sample is weighted equally. 
When $\eta \to 1$, each sample is reweighted in terms of its class frequency. 
To take advantage of resampling and reweighting, we first resample transitions from the dataset and then reweight these samples during training.

\subsection{Uncertainty Correction}
The proposed resampling and reweighting might assign more importance to sparse samples with higher uncertainty compared with that of the majority classes. 
When combined with function approximation, the state-action pairs near the minority samples might be assigned with excessive values, which will further amplify the extrapolation error. 
To overcome this issue, we restrain state-action pairs, which are out-of-distribution near the minority classes, more strictly.
Our method of dealing with data imbalance can be generally integrated into prior methods.
For policy constrained methods\cite{kumar2019stabilizing}, the threshold of the Lagrange multiplier could be reweighted by the inverse of the class frequency. In this manner, the agent is forced to keep closer to the actions in the batch when it is located in uncertain regions. For conservative Q-learning, the minimization and maximization terms could be weighted with the inverse of the class frequency as well.
The update rule of uncertainty correction is defined as:

\begin{equation}
\begin{aligned}   
\min _{Q}  \quad & \alpha \cdot (\mathbb{E}_{s \sim \mathcal{D}, a \sim \pi(a \mid s)}[w(s,a)Q(s, a)] -
\\ & \mathbb{E}_{s \sim \mathcal{D}, a \sim \hat{\pi}_{\beta}(a \mid s)}[w(s,a)Q(s, a)] ) \\ & +\frac{1}{2} \mathbb{E}_{s,a, s^{\prime} \sim \mathcal{D}}\left[w(s,a)\left(Q(s, a)-\hat{\mathcal{T}}^{\pi} \hat{Q}(s, a)\right)^{2}\right]. 
\end{aligned}
\label{eq6.4}
\end{equation}

\subsection{Implementations}
Our proposed method for data imbalance issues could be employed as a general framework for any offline reinforcement method.
In this paper, we implement our proposed method on the basis of CQL.
We integrate the tighter lower bound into the proposed framework. 
In order to estimate the Pearson $\chi^2$ diverge between $\pi_\beta(a|s)$ and $\pi(a|s)$, the lower bound can be obtained using function approximation as:
\begin{equation}
    F(\vartheta)=\mathbb{E}_{a \sim \pi_\beta(a|s)}\left[g_f(T_{\vartheta}(a))\right]-\mathbb{E}_{a \sim \pi(a|s)}\left[f^{*}\left(g_f(T_{\vartheta}(a))\right)\right],
    \label{eq:pearson}
\end{equation}
where $g_f(u) = u$.
In Eq. \ref{eq:pearson}, $f^*(t)=\frac{1}{4}t^2+t$ denotes Pearson $\chi^2$ divergence and $\vartheta$ is the parameter of deep neural network.
During the experiment, we observed that setting a hyperparameter $\delta$ is 
a satisfactory alternative to stabilize the training process. 
Therefore, the objective of our proposed method is finally defined as:
\begin{equation}
\begin{aligned}   \min _{Q}  \quad & \alpha \cdot (\mathbb{E}_{s \sim \mathcal{D}, a \sim \pi(a \mid s)}[w(s,a)Q(s, a)] -
\\ & \mathbb{E}_{s \sim \mathcal{D}, a \sim \hat{\pi}_{\beta}(a \mid s)}[w(s,a)(1+\delta)Q(s, a)] ) \\ & +\frac{1}{2} \mathbb{E}_{s,a, s^{\prime} \sim \mathcal{D}}\left[w(s,a)\left(Q(s, a)-\hat{\mathcal{T}}^{\pi} \hat{Q}(s, a)\right)^{2}\right] \end{aligned}
\label{eq6.6}
\end{equation}
The whole process of our proposed method is summarized in Algorithm \ref{alg:algorithm}. 

\begin{algorithm}
\caption{}
\label{alg:algorithm}
\textbf{Input}:  offline dataset $\mathcal{D}$, update iterations $t_{max}$ for policy and Q-value networks, Pearson $\chi^2$ divergence hyperparameter $\delta$, number of clusters $k$\\
\textbf{Parameter}: policy network $\pi$, Q-networks $Q_1, Q_2$. \\
\textbf{Output}: learnt policy network $\pi$
\begin{algorithmic}[1] 
\STATE Initialize the policy network, Q-networks.
\STATE Conduct k-means on the dataset until convergence
\STATE Calculate cluster frequency of each cluster.
\STATE Compute the weight of each cluster according to the cluster frequency.

\WHILE{$ t < t_{max}$}
\STATE Resample a mini-batch of samples from $\mathcal{D}$ according to Eq. \ref{eq6.1}.
\STATE Calculate weights according to Eq. \ref{eq6.3}.
\STATE Update the Q-networks according to Eq.\ref{eq6.6}.
\STATE Update the policy network.
\ENDWHILE

\end{algorithmic}
\end{algorithm}

\section{Experiments}

\subsection{Clustering Performance}
The performances of resampling and reweighting are directly related to the k-means clustering. 
We first study the effectiveness of the clustering method. 
We choose Halfcheetah and Walker2d tasks for evaluation.
Different policies are utilized to generate corresponding samples and form a mixed dataset.
The dataset consists of $900,000$ ``random'' samples generated by rollouting a random policy, and $100,000$ ``expert'' samples generated by a well trained soft actor-critic policy.

We perform k-means clustering with the cluster number as $2$. 
Since the k-means clustering is an unsupervised learning method, we associate the cluster with the majority of all samples as cluster 1 and the rest of the samples as cluster 2. 
As shown in Table \ref{tb1}, cluster 1 contains most of the ``random'' samples while cluster 2 contains most of the ``expert'' samples and a small number of ``random'' samples.

\begin{figure}[ht]
\centering  
\subfigure[Halfcheetah]{\includegraphics[width=0.47\linewidth]{ 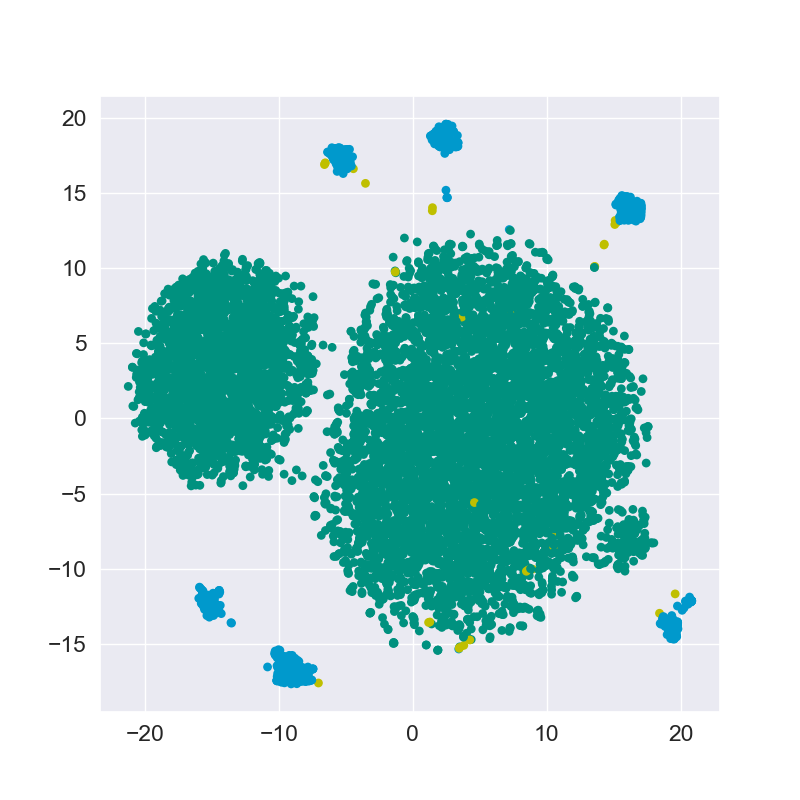}}
\subfigure[Walker2d]{\includegraphics[width=0.47\linewidth]{ 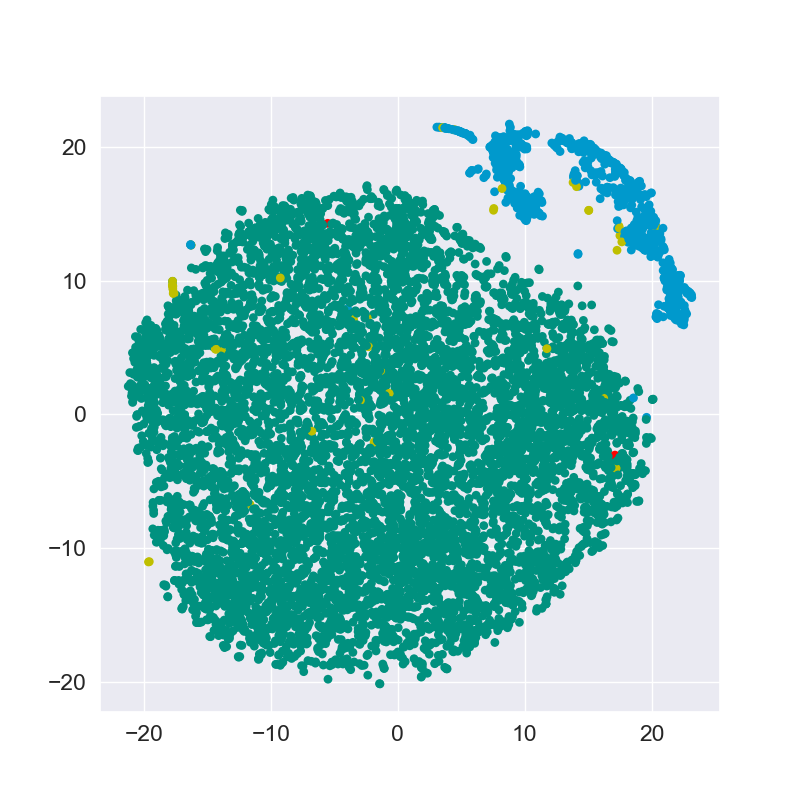}}
\caption{Data distribution of samples obtained on mixed Halfcheetah and Walker2d tasks using k means clustering.}
\label{fig7.1:kmeans}
\end{figure}

We further use t-Distributed Stochastic Neighbor Embedding (t-SNE) to visualize the data distribution. 
The visualization for Walker2d and Halfcheetah tasks are shown in Fig. \ref{fig7.1:kmeans}.
In Fig. \ref{fig7.1:kmeans}, the green and red points denote the ``random'' and ``expert'' samples in cluster 1 respectively, while the yellow and blue points denote ``random'' and ``expert'' samples in cluster 2 respectively.  
Note that red points are so rare and hard to distinguish.
The experimental results show that ``random'' samples and ``expert'' samples are able to be separated by k-means clustering.

\begin{table}[bh]
\centering
\begin{tabular}{l|cccccccc}
    \toprule
Task &  Data & cluster 1  & cluster 2  \\
    \midrule

Halfcheetah &   random  &  899100 & 0  \\
            &   expert  &  4553    & 95346 \\
    \midrule

Walker2d    &   random  &  899889 & 110  \\
            &   expert  &  4298    & 95634 \\       
    \bottomrule
\end{tabular}
\caption{The number of samples in different clusters for Halfcheetah and Walker2d.}
\label{tb1}
\end{table}

\subsection{Offline Mujoco Control Datasets}
\begin{figure}[ht] 
\centering 
\subfigure[Halfcheetah]{\includegraphics[width=0.3\linewidth, height=0.2\linewidth]{ 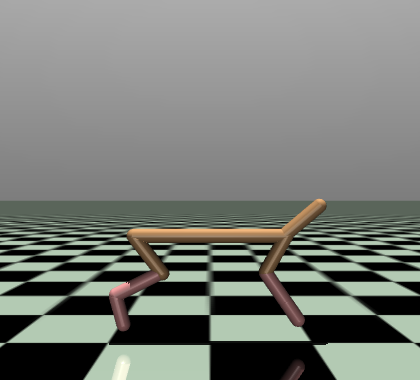}}
\subfigure[Hopper]{\includegraphics[width=0.3\linewidth, height=0.2\linewidth]{ 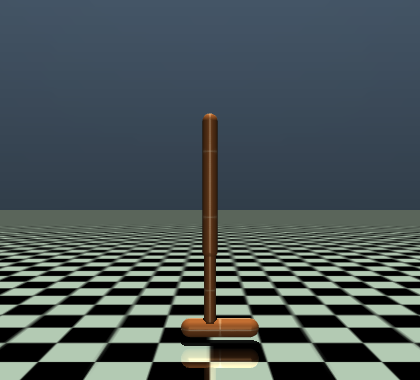}}
\subfigure[Walker2d]{\includegraphics[width=0.3\linewidth, height=0.2\linewidth]{ 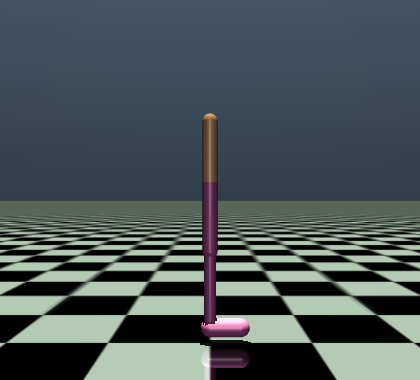}}
\caption{The Mujoco environments.}
\label{fig7.2:env}
\end{figure}

To evaluate the effectiveness of our proposed method, we perform experiments using the Mujoco control suits in D4RL benchmarks \cite{fu2020d4rl}. 
D4RL benchmarks present static datasets for Hopper, Halfcheetah, and Walker2d Mujoco tasks (as shown in Fig. \ref{fig7.2:env}). 
For each type of control environment, D4RL offers $5$ kinds of datasets, ``random'', ``medium'', ``medium-replay'', ``medium-expert'', and ``expert'' dataset. 
The ``expert'' dataset consists of $10^6$ samples generated by a well trained soft actor-critic policy. 
The ``medium'' dataset is generated by a policy which is trained to achieve approximate $1/3$ performance as that of the ``expert'' dataset. 
The ``random'' dataset is generated by rollouting a randomly initialized policy for $10^6$ steps. 
The ``medium-replay'' dataset contains all samples stored in the replay buffer during training policy for the ``medium'' dataset.
The ``medium-expert'' dataset contains the same number of samples selected from the ``medium'' and the ``expert'' dataset.

\begin{figure}[h] 
\centering  
\includegraphics[width=1.0\linewidth]{ 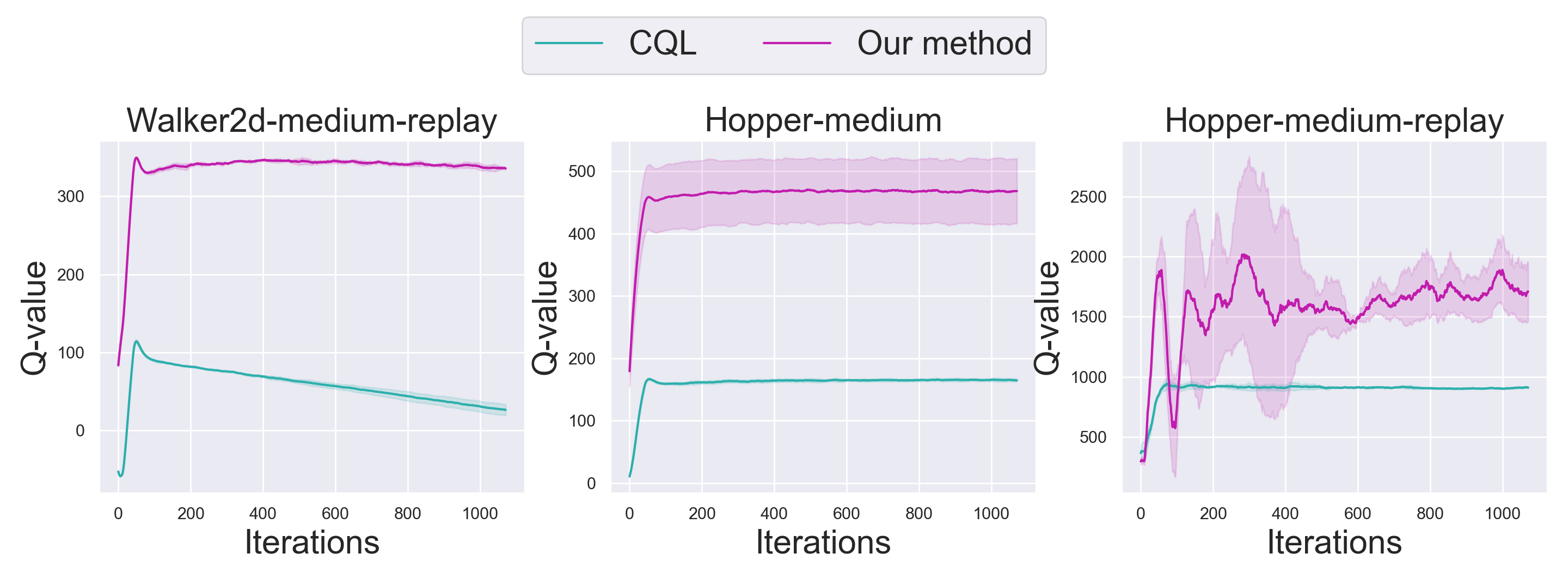}
\caption{Q-value of our method compared to CQL on ``Walker2d-medium-replay'', ``Hopper-medium'', and ``Hopper-medium-replay''.}
\label{fig:qvalue}
\end{figure}

The experimental results of the state-of-the-art methods and our proposed methods are summarized in Table \ref{tb1:mujoco}. 
The results for BEAR\cite{kumar2019stabilizing}, BRAC\cite{wu2019behavior}, SAC\cite{haarnoja2018soft}, and BC are presented by \cite{fu2020d4rl}, and the result of CQL is presented by \cite{kumar2020conservative}. 
As shown in Table \ref{tb1:mujoco}, our proposed method outperforms or achieves competitive performance compared the state-of-the-art methods on most tasks.
Especially, our proposed method exceeds the best previous methods on the mixed ``medium' and ``medium-replay'' datasets. 
Our proposed method achieves more than $3,000$ scores on the ``Hopper-medium'' task. As shown in Fig. \ref{fig:qvalue}, our method learns a higher Q-value than CQL.
The experiments demonstrate that the proposed reducing conservativeness oriented offline reinforcement learning is helpful to non-expert tasks and beneficial for tackling data imbalance issues. 

\begin{table*}[ht]
\centering
\begin{tabular}{l|cccccccc}
    \toprule
Task Name & SAC & BC & BEAR & BRAC-p & BRAC-v & CQL  & Our method\\
    \midrule
    Halfcheetah-random &30.5& 2.1 & 25.5 & 23.5 & 28.1 & \textbf{35.4}    & \textbf{35.3}\\
    Walker2d-random &4.1& 1.6 & 6.7 & 0.8 & 0.5 & 7.0          & 7.2\\
    Hopper-random &11.3& 9.8 & 9.5 & 11.1 & 12.0 & 10.8   & 10.7\\

    Halfcheetah-medium &-4.3& 36.1 & 38.6 & 44.0 & 45.4 & 44.4   & \textbf{51.4} \\
    Walker2d-medium &0.9& 6.6 & 33.2 & 72.7 & \textbf{81.3} & 79.2  & 78.8 \\
    Hopper-medium &0.8& 29.0 & 47.6 & 31.2 & 32.3 & 58.0   & \textbf{96.8} \\
    
        Halfcheetah-medium-replay &-2.4& 38.4 & 36.2 & 45.6 & 46.9 & 46.2  & \textbf{51.5} \\
    Walker2d-medium-replay &1.9& 11.3 & 10.8 & -0.3 & 0.9 & 26.7   & \textbf{46.2}\\
    Hopper-medium-replay & 3.5& 11.8 & 25.3 & 0.7 & 0.8 & 48.6  & \textbf{65.2} \\
    
        Halfcheetah-medium-expert &1.8 & 35.8 & 51.7 & 43.8 & 45.3 & \textbf{62.4}   & 60.3 \\
    Walker2d-medium-expert & 1.9   & 11.3 & 10.8 & -0.3 & 0.9 & 98.7   & \textbf{105.3} \\
    Hopper-medium-expert & 1.6 & 111.9 & 4.0 & 1.1 & 0.8 & \textbf{111.0}   & \textbf{110.9} \\
    
        Halfcheetah-expert &-1.9& \textbf{107.0} & \textbf{108.2} & 3.8 & -1.1 & 104.8   & 102.6\\
    Walker2d-expert &-0.3& 125.7 & 106.1 & -0.2 & -0.0 & \textbf{153.9}  & 108.3 \\
    Hopper-expert &0.7& \textbf{109.0} & \textbf{110.3} & 6.6 & 3.7 & \textbf{109.9}  & \textbf{111.6}\\
    
    \bottomrule
\end{tabular}
\caption{Performance of SAC, BC, BEAR, BRAC, CQL, and our proposed method on the offline Mujoco control suite tasks. The results are obtained using averaged over four seeds with the normalized return metric. Note that our proposed method achieves competitive performance compared with that of the methods on most tasks.}
\label{tb1:mujoco}
\end{table*}

\subsection{Mixed Datasets}
\begin{figure}[th] 
\centering  
\includegraphics[width=1.0\linewidth]{ 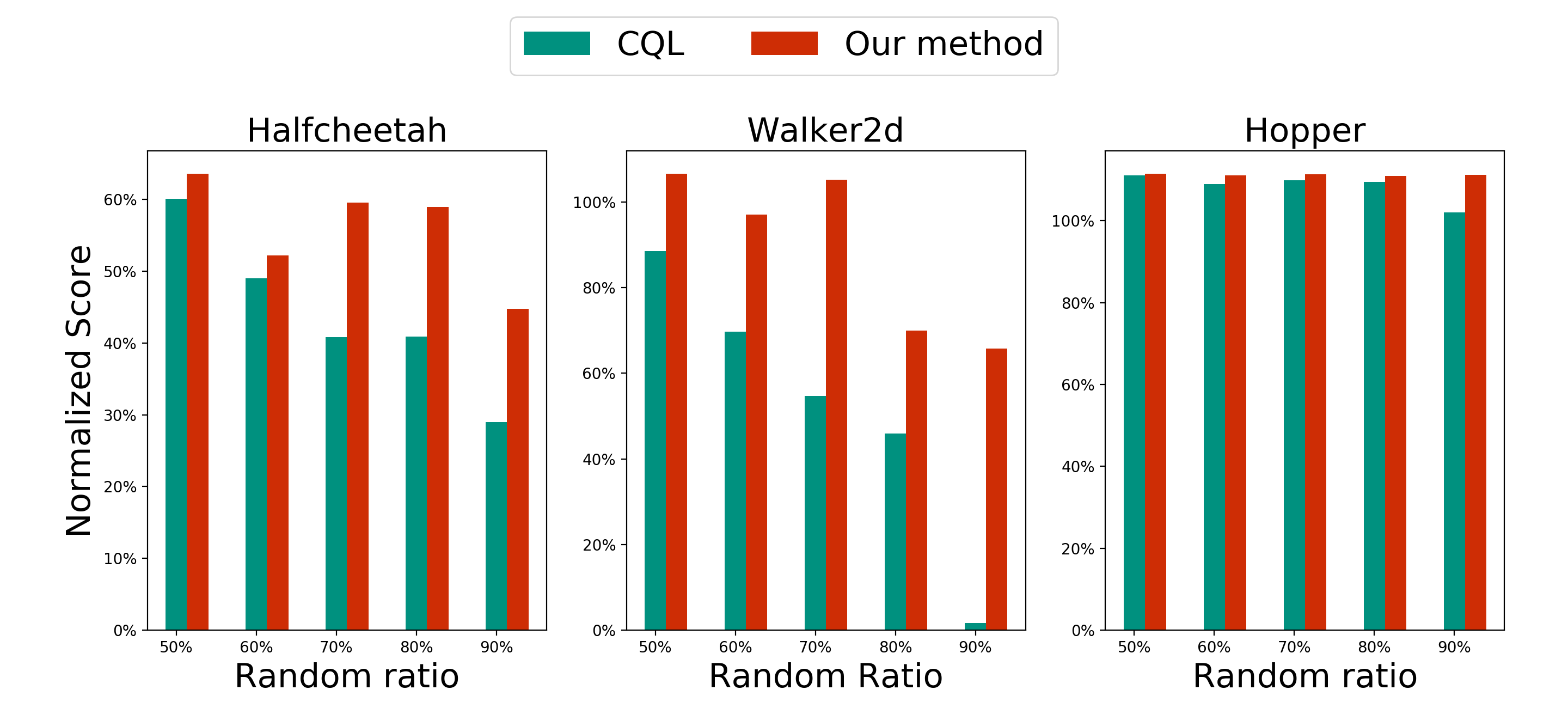}
\caption{Results of CQL and our proposed method on the mixed dataset. The results are obtained using averaged over four seeds with on the normalized return metric. Note that our method outperforms CQL greatly on mixed Halfcheetah and Walker2d tasks.}
\label{fig7.3:imbalance}
\end{figure}

We further evaluate the ability of our proposed method for tackling data imbalance issues on our own designed mixed dataset.
We create $5$ datasets for each of the Halfcheetah, Walker2d, and Hopper tasks.
Each dataset consists of $10^6$ samples with mixed ``random'' and ``expert'' samples.
For each task, the proportions of random samples in the $5$ datasets are $0.5$, $0.6$, $0.7$, $0.8$, and $0.9$ respectively.
We compare our proposed method with CQL and show the results in Fig. \ref{fig7.3:imbalance}. 
As expected, using more ``random '' samples will affects the performance of the policy.

As shown in Fig. \ref{fig7.3:imbalance}, our proposed method considerably outperforms CQL on most of the tasks.  
For Halfcheetah and Walker2d tasks, the performance of our proposed method decreases much more slowly. 
For Hopper tasks, our proposed method is competitive to CQL in the $4$ tasks and outperforms CQL with a small margin when the proportion of the ``random'' samples is set as $0.9$. 
The improvements obtained compared with that of CQL demonstrate that our proposed method is less sensitive to the reduced ``expert'' samples, and it can utilize sparse samples more effectively.

\subsection{Ablation Study}
In our proposed method, the number of clusters and the choice of $\delta$ are important hyperparameters.
In this section, we present the experiments for the ablation study.

\subsubsection{Number of clusters}
\begin{figure}[ht] 
\centering  
\includegraphics[width=1.0\linewidth, height=0.5\linewidth]{ 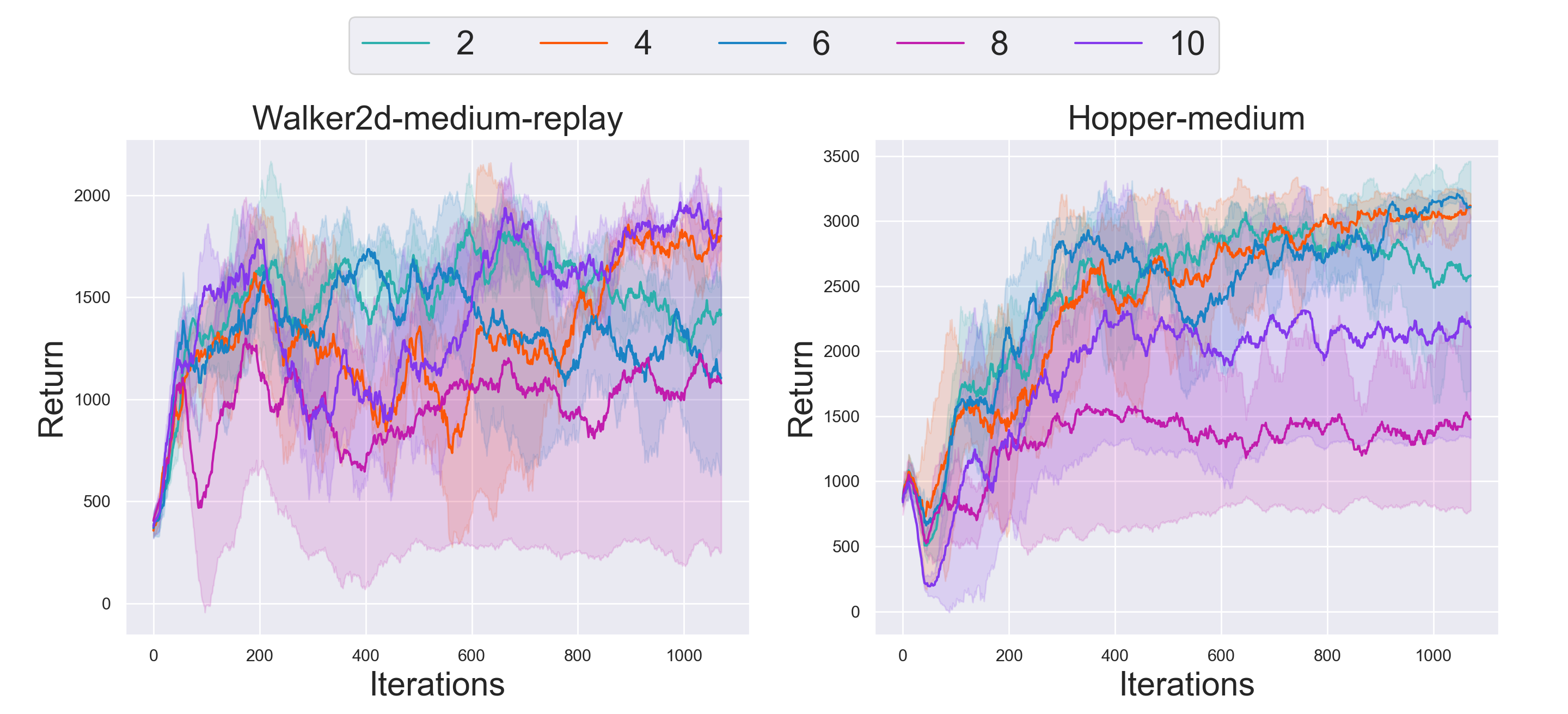}
\caption{Results of our proposed method on ``Walker2d-medium-replay'' and ``Hopper-medium'' tasks using different $k$. }
\label{fig7.4:cluster}
\end{figure}

We evaluate the sensitivity of our  proposed method using different numbers of clusters. 
We perform the experiments on ``Walker2d-medium-replay'' and ``Hopper-medium'' tasks by using cluster number $k$ as $2$, $4$, $6$, $8$, and $10$.
The experimental results are depicted in Fig. \ref{fig7.4:cluster}.

As shown in Fig. \ref{fig7.4:cluster}, the performance of our proposed method first increases and then decreases as $k$ increases. 
Due to the high dimensional state-action space, the samples are difficult to be separated.
One promising solution is to utilize more concrete representation to formulate state-action samples.

\subsubsection{Choice of $\delta$}
\begin{figure}[ht] 
\centering  
\includegraphics[width=1.0\linewidth, height=0.5\linewidth]{ 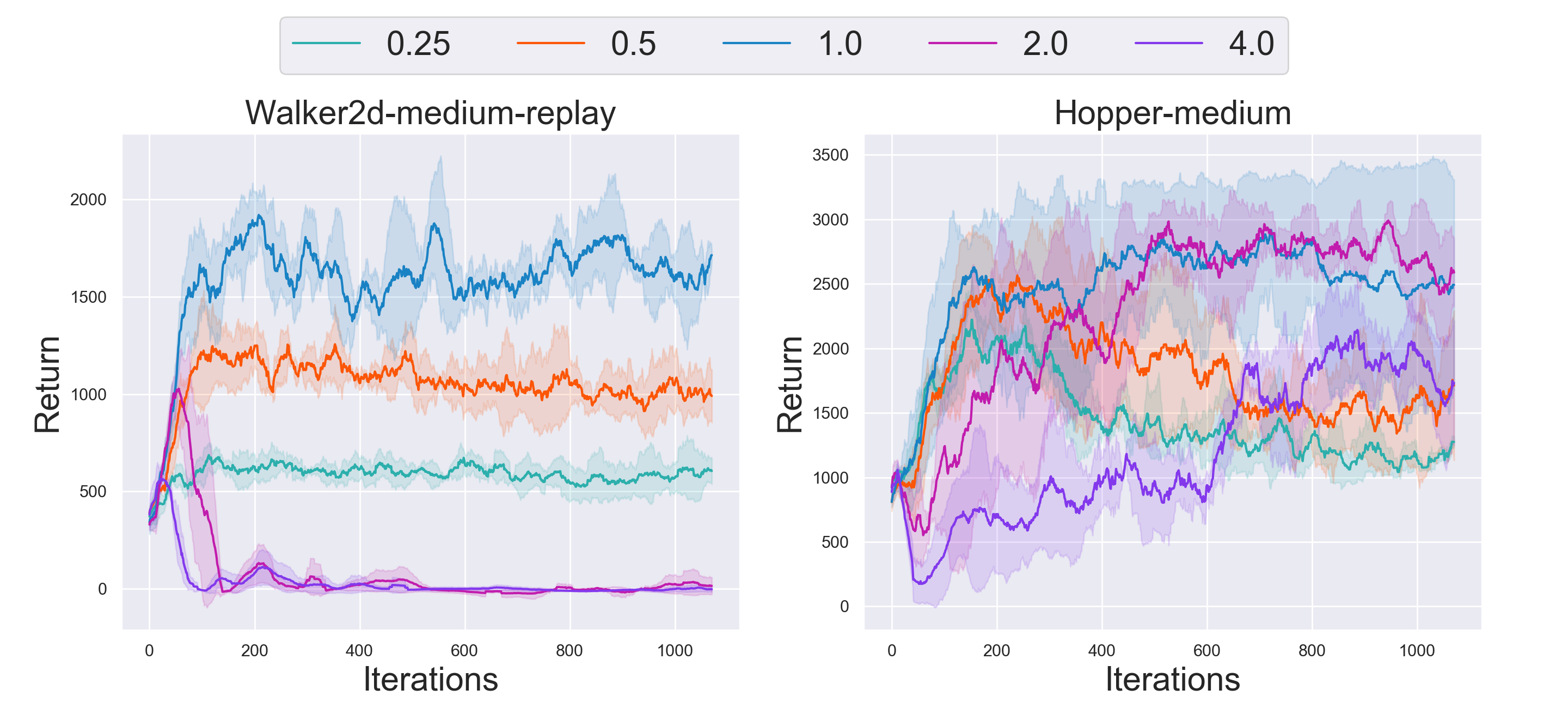}
\caption{Results of our proposed method on ``Walker2d-medium-replay'' and ``Hopper-medium'' tasks using different $\delta$.}
\label{fig7.5:delta}
\end{figure}

In our proposed method, $\delta$ is a critical hyperparameter for controlling the degree of conservativeness. 
We evaluate the performance by setting $\delta$ as $0.25$, $0.5$, $1.0$, $2.0$, and $4.0$, and perform the experiments on ``Walker2d-medium-replay'' and ``Hopper-medium'' tasks. 
As shown in Fig. \ref{fig7.5:delta}, when $\delta$ is set with a large value, the Q-value function tends to increase to infinity and the accumulative reward tends to be zero. 
When $\delta$ is set with a small value, the proposed method becomes very conservative and yields poor performance.

\section{Conclusions}
In this paper, we introduce the method of reducing conservativeness oriented offline reinforcement learning. 
After partitioning the samples into different clusters, we reweight and resample them according to their class frequency. To reduce extrapolation error, we put more constraints on those minority classes. Besides, we introduce a tighter lower bound of the value function. Our experimental results show that our method outperforms state-of-art approaches on D4RL evaluation benchmarks and our designed mixed datasets.

%\clearpage

\bibliography{main}
\bibliographystyle{icml2021}

\end{document}